\documentclass[a4paper]{ceurart}

\usepackage{algorithmic}
\usepackage{graphicx}
\usepackage{textcomp}
\usepackage{xcolor}
\usepackage{booktabs}
\usepackage{array}
\usepackage{longtable}
\usepackage{capt-of}
\usepackage{float}
\usepackage{soul}

\sethlcolor{yellow}
\newcommand{\new}[1]{#1}

\begin{document}

\copyrightyear{2026}
\copyrightclause{Copyright for this paper by its authors.
  Use permitted under Creative Commons License Attribution 4.0
  International (CC BY 4.0).}

\conference{CLEAR-AI 2026: Workshop on Collaborative Methods and Tools for
  Engineering and Evaluating Transparency in AI, co-located with IJCAI-ECAI 2026,
  August 16, 2026, Bremen, Germany}

\title{DeBERTa-Sentinel: Toward Transparent and Trustworthy Detection of AI-Generated Text}

\author[1]{Muhammad Yousaf Rehman}
\fnmark[1]

\author[2,3]{Muhammad Islam} [orcid=0000-0002-4845-4808]
\cormark[1]
\fnmark[1]

\address[1]{SPECS, University of Hertfordshire, Hatfield, UK}
\address[2]{College of Science and Engineering, James Cook University, Cairns, 4878, QLD, Australia}
\address[3]{Centre for AI and Data Science Innovation, James Cook University, Cairns, QLD 4878, Australia}

\cortext[1]{Corresponding author.}
\fntext[1]{These authors contributed equally to this work.}

\begin{abstract}
The rapid spread of large language models (LLMs) across the web raises concerns about misinformation, academic integrity, automated content manipulation, and risks to vulnerable online communities. Existing transformer-based detectors, such as GPT-Sentinel, show promise but struggle to generalize to diverse model outputs and paraphrasing attacks, limiting their role in building trustworthy web ecosystems. This work introduces DeBERTa-Sentinel, a responsible AI-generated text detection framework leveraging DeBERTa-v3's disentangled attention to capture subtle structural irregularities in synthetic content. A central design principle is transparency: unlike black-box commercial detectors, DeBERTa-Sentinel exposes token-level explanations of its decisions, enabling affected stakeholders journalists, educators, and platform trust and safety teams to audit, challenge, and contextualize detection outcomes. Using the GLC-AIText dataset of 28,057 human and LLM-generated samples (GPT, LLaMA, and Claude) with a 60-20-20 split, DeBERTa-Sentinel achieves 98.21\% validation accuracy and surpasses the RoBERTa-Sentinel baseline from NeurIPS 2025, achieving 97.53\% test accuracy, 95.89\% precision, 99.33\% recall, and 99.53\% ROC-AUC, and maintaining a 0.665\% false negative rate. The model's interpretability reveals linguistic markers such as academic phrasing and formal transitions associated with synthetic text, directly supporting stakeholder needs for verifiable, auditable content-authenticity decisions. By advancing responsible detection methods that reduce bias and enhance explainability, DeBERTa-Sentinel promotes trustworthy, ethical, and human-centric AI systems. Code and data are available at \url{https://github.com/Galileo-Galili/HUMAN-VS-AI-TEXT-DETECTION}.
\end{abstract}


\maketitle

\section{Introduction}
As large language models (LLMs) like GPT \cite{openai2023gpt35}, LLaMA \cite{touvron2023llama}, and Claude \cite{anthropic2023claude} grow increasingly capable, distinguishing AI-generated text from human writing has become a critical challenge across journalism, education, and legal domains. While early approaches relied on statistical methods like logistic regression and SVMs, the emergence of advanced LLMs necessitated the use of deep transformer-based detectors \new{\mbox{\cite{wu2025survey}}}. \new{Transformer-based classifiers have likewise become the dominant approach for related web-content integrity tasks such as misinformation and fake news detection, including in low-resource language settings \mbox{\cite{islam2025unified}}.}

One such method, GPT-Sentinel \cite{chen2023gpt}, used a frozen RoBERTa encoder with a classification head, achieving strong results on GPT-generated text. However, recent studies have revealed key limitations in such detectors---particularly their lack of generalization to outputs from diverse models and paraphrased inputs \cite{trivedi2025defactify, krishna2023paraphrasing}. These limitations stem partly from architectural constraints inherent to RoBERTa: its attention mechanism processes content and positional information jointly within a single representation space. This entanglement means that when the model computes attention scores between tokens, it cannot separately evaluate ``what is being said'' (semantic content) versus ``where it appears'' (structural position). Consequently, RoBERTa may struggle to detect subtle positional regularities---such as the consistent placement of hedging phrases or the formulaic positioning of topic sentences---that are characteristic markers of synthetic text.

\new{Beyond raw detection accuracy, transparency has become an increasingly important consideration in AI-generated text detection, with explainable methods providing insight into the linguistic and stylistic features that influence detector predictions \mbox{\cite{shah2023detecting}}. In practical deployments, transparent and interpretable outputs are valuable for stakeholders such as educators investigating potential academic misconduct, journalists assessing the authenticity of written content, and platform moderators enforcing content policies, as they provide supporting evidence that can be reviewed alongside human judgment rather than relying solely on opaque model predictions. This need is further reinforced by evidence that existing detectors can exhibit systematic biases against non-native English writers, highlighting the importance of decisions that can be inspected and challenged when necessary \mbox{\cite{liang2023gpt}}.}

In this study, we improve upon the GPT-Sentinel framework by introducing DeBERTa-Sentinel, a classifier that replaces the RoBERTa backbone with DeBERTa-v3. DeBERTa addresses these limitations through disentangled attention and enhanced relative positional encoding \cite{he2021deberta}, which we describe fully in Section~\ref{sec:methodology}. We also diversify the training data by incorporating outputs from multiple LLMs (GPT-3.5, LLaMA, Claude), improving robustness across generation styles. \new{Crucially, DeBERTa-Sentinel is designed from the outset for transparency: it provides token-level attribution that allows end-users to understand and audit which textual features drive each detection decision.}

Our contributions are fourfold:
\begin{enumerate}
\item We present DeBERTa-Sentinel, a novel detection framework that incorporates disentangled attention to more effectively capture subtle structural irregularities in synthetic text\new{, while providing built-in token-level explainability for transparent, auditable predictions}.
\item We construct the GLC-AIText dataset, consisting of 28,057 paraphrased samples produced by diverse large language models (GPT-3.5, LLaMA, and Claude), with the aim of enabling broader generalization and improving model robustness.
\item Our empirical evaluation demonstrates that DeBERTa-Sentinel achieves consistent improvements over RoBERTa-Sentinel and traditional machine learning baselines, including a held-out generator experiment where the model achieves 98.46\% accuracy on Claude-generated text despite being trained exclusively on GPT-3.5 and LLaMA outputs, demonstrating strong cross-generator generalization.
\item An explainability analysis shows the proposed model provides interpretable token-level predictions, identifying discriminative linguistic patterns such as transitional phrases and academic terminology commonly induced in synthetic content\new{---directly enabling the transparency requirements of journalists, educators, and civic institutions}.
\end{enumerate}

The rest of this paper is structured as follows: Section~\ref{sec:related} surveys related work; Section~\ref{sec:dataset} details our dataset; Section~\ref{sec:methodology} outlines the model architecture and training; Section~\ref{sec:evaluation} presents evaluation results; \new{Section~\mbox{\ref{sec:transparency}} discusses transparency and responsible deployment;} Section~\ref{sec:discussion} provides further discussion; and Section~\ref{sec:conclusion} concludes the article.

\section{Related Work}
\label{sec:related}
The detection of AI-generated text has emerged as a critical area of research, particularly with the rapid proliferation of large language models (LLMs) capable of producing highly fluent and coherent text \cite{generativeAI}\new{\mbox{\cite{wu2025survey}}}. One of the earliest prominent contributions in this domain was GLTR \cite{gehrmann2019gltr}, which employed statistical analysis of token-level probabilities and rank-based detection methods to visualize and identify machine-generated content. GLTR used probability distribution analysis and entropy calculations to detect synthetic text, but suffered from limited adaptability to evolving generation techniques and poor performance against newer language models.

Subsequent works, such as Ippolito et al.~\cite{ippolito2020automatic}, utilized perplexity-based detection methods and human evaluation studies, showing that automatic detectors perform best when generated text closely resembles human-authored text. However, these approaches were limited by their reliance on surface-level linguistic features and poor generalization across different domains. Similarly, Solaiman et al.~\cite{solaiman2019release} employed content-based filtering and release strategy frameworks but was constrained by the lack of robust detection mechanisms for sophisticated language models.

A more recent advancement is GPT-Sentinel \cite{chen2023gpt}, which introduced a transformer-based classification framework using frozen RoBERTa encoders with a multi-layer perceptron (MLP) classification head. GPT-Sentinel demonstrated strong performance on binary classification between human and AI-generated text, particularly from GPT-3.5, but was limited by RoBERTa's joint processing of content and positional information, which can obscure the structural patterns that distinguish synthetic text, and poor generalization to newer generative models.

To overcome these architectural limitations, the present study draws on DeBERTa \cite{he2021deberta}, which incorporates disentangled attention mechanisms and enhanced relative positional encoding. Through mathematical decomposition of the attention mechanism, DeBERTa independently models content-to-content, content-to-position, and position-to-content interactions, enabling more precise identification of structural artifacts in synthetic text. While existing implementations have demonstrated DeBERTa's effectiveness across various NLP tasks, limited work has explored task-specific optimization of its disentangled architecture for AI text detection.

Further comparative evaluations by Trivedi and Sivanesan \cite{trivedi2025defactify} demonstrated ensemble methods with DeBERTa-based detectors trained on model-diverse data, achieving improved robustness\new{. Complementary work has emphasized interpretability in detection: Shah et al.~\mbox{\cite{shah2023detecting}} combine stylistic features with explainable AI techniques to detect and unmask AI-generated text, reinforcing that interpretability is now a first-class design concern alongside raw accuracy.}

Recent research has explored zero-shot detection methods, with Mitchell et al.~\cite{mitchell2023detectgpt} introducing DetectGPT, which uses probability curvature analysis for machine-generated text detection without requiring training data. However, Krishna et al.~\cite{krishna2023paraphrasing} demonstrated that paraphrasing techniques can evade many detection systems, highlighting vulnerabilities in current approaches and the need for more robust detection frameworks.

The challenge of reliable detection has been further emphasized by studies examining the limitations of commercial detection tools \cite{weber2023public, sadasivan2023can}, which demonstrated that existing detectors show bias against non-native English writers \cite{liang2023gpt} and struggle with adversarial modifications to AI-generated content. \new{Studies of commercial AI-text detection tools have shown that their performance can be unreliable in educational settings and that some detectors exhibit systematic bias against non-native English writers, raising concerns about their use in high-stakes academic assessment\mbox{\cite{weber2023public,liang2023gpt}}.}

In summary, while prior works established foundational detection approaches, they often fail to capture deeper representational differences between human and synthetic text, especially as LLMs become more advanced. By introducing DeBERTa's disentangled attention into the classification pipeline\new{, curating a more diverse synthetic corpus, and embedding token-level explainability as a core design feature}, our study addresses these architectural\new{, methodological, and transparency} limitations.

\section{Dataset}
\label{sec:dataset}

This study builds upon the \textit{OpenGPTText} dataset introduced by \cite{chen2023gpt}, which paired human-written samples from OpenWebText with paraphrased output from OpenAI's gpt-3.5-turbo model. To improve generalization and robustness in AI-generated text detection, we extend the dataset by generating paraphrases using multiple LLMs while keeping the original human-written samples unchanged.

\subsection{GLC-AIText Overview}
The GLC-AIText dataset consists of paraphrased textual samples that were generated using a combination of gpt-3.5-turbo, Meta's LLaMA, and Claude language models. The original human-written content was taken from the cleaned OpenWebText corpus, specifically the OpenWebText-Final subset as used in the base paper \cite{chen2023gpt}, and serves as the shared source for paraphrasing across all models. Each paraphrased sample corresponds to a human-written text from the OpenWebText corpus and shares the same unique identifier (UID) to preserve alignment.

The GLC-AIText dataset contains approximately 28,057 paraphrased samples, covering around 1\% of the full OpenWebText corpus in specific subsets. The same subset IDs were used to collect both human-written and paraphrased samples. The number of samples in each subset is listed in Table~\ref{tab:dataset_breakdown}.

\begin{table}[t]
\centering
\caption{Dataset breakdown by subset and model}
\label{tab:dataset_breakdown}
\footnotesize
\begin{tabular}{lccc}
\toprule
\textbf{Subset} & \textbf{GPT Generated} & \textbf{LLaMA Generated} & \textbf{Claude Generated} \\
\midrule
Urlsf\_00 & 1386 & 1300 & 1212 \\
Urlsf\_01 & 1308 & 1300 & 801 \\
Urlsf\_02 & 874 & 1300 & 922 \\
Urlsf\_03 & 1299 & 1300 & 1295 \\
Urlsf\_04 & 1156 & 1300 & 946 \\
Urlsf\_05 & 1103 & 1300 & 1164 \\
Urlsf\_06 & 1031 & 1300 & 723 \\
Urlsf\_09 & 1217 & 1300 & 1220 \\
\midrule
\textbf{Total} & \textbf{9374} & \textbf{10400} & \textbf{8283} \\
\bottomrule
\end{tabular}
\end{table}

\subsection{Data Source}
The human-written samples used in this study were obtained from the OpenWebText corpus, a publicly available dataset comprising web content sourced from URLs shared on Reddit with a minimum of three upvotes \cite{gokaslan2019openwebtext}. This corpus serves as a reconstitution of the original WebText dataset described by \cite{radford2019language}. Since the OpenWebText corpus was compiled in 2019, the textual content it contains was not algorithmically generated, making it suitable as ground truth human-written text for AI detection tasks.

Specifically, we utilized the cleaned OpenWebText-Final subset provided by the GPT-Sentinel dataset \cite{chen2023gpt}, which underwent preprocessing to remove stylistic disparities and normalize formatting inconsistencies. This cleaned version ensures that our models focus on semantic and linguistic features rather than superficial formatting artifacts that could lead to overfitting.

\subsection{Data Collection Method}
We constructed the AI-generated portion by paraphrasing the cleaned human-written samples from the OpenWebText-Final corpus \cite{chen2023gpt}, covering subsets urlsf\_00 to urlsf\_06 and urlsf\_09. Each subset was divided into three parts: one retained gpt-3.5-turbo paraphrases from the base dataset, while the other two were paraphrased using Meta's LLaMA and Claude models.

All models were prompted with the same instruction used in the original OpenGPTText collection: ``Rephrase the following paragraph by paragraph.'' Following the methodology of \cite{chen2023gpt}, samples longer than 2,000 words were filtered out due to model input limitations, and content blocked by safety filters was excluded. Outputs were additionally filtered for fluency and coherence, and near-duplicate generations were removed.

Both the original human-written OpenWebText samples and the AI-generated paraphrases from all three models (GPT-3.5, LLaMA, and Claude) were used during training and testing phases to ensure comprehensive exposure to diverse writing styles and generation patterns.

\subsection{\new{Choice of Generator Models}}
\label{sec:model_choice}

\new{
Rather than more recent frontier models such as GPT-4o or Claude~3.5, we deliberately selected GPT-3.5, LLaMA, and Claude for three reasons. First, they span architecturally distinct LLM families, a proprietary, RLHF-tuned chat model from OpenAI \mbox{\cite{brown2020language,ouyang2022instructgpt}}, Meta's open-weight foundation model \mbox{\cite{touvron2023llama}}, and Anthropic's Constitutional AI model \mbox{\cite{bai2022constitutional}}, so training on their outputs exposes the detector to diverse generation styles and reduces dependence on single-generator artifacts. Second, robust detectors should generalize across generators and domains: the MAGE benchmark shows that existing detectors degrade sharply on unseen generators \mbox{\cite{li2024mage}}, and our use of \emph{paraphrased} text, which strongly evades detection \mbox{\cite{krishna2023paraphrasing}}, further stresses this setting. Third, GPT-3.5 enables a fair comparison with GPT-Sentinel, while holding out Claude (Section~\mbox{\ref{sec:heldout}}) tests whether the detector learns transferable rather than generator-specific representations.
}

\subsection{Dataset Cleaning}
The human-written samples from OpenWebText-Final had already undergone cleaning in the base paper \cite{chen2023gpt}, including removal of excessive newline characters and mapping of Unicode characters to ASCII equivalents. This preprocessing was originally implemented to eliminate potential confounding factors such as ChatGPT's tendency to use Unicode quotation marks (U+201D) instead of ASCII quotation marks (U+0022) and different newline formatting patterns.

For the AI-generated paraphrases from LLaMA and Claude, we retained their raw output form with minimal post-processing, applying only basic filtering to remove incoherent or incomplete generations. Unlike the base paper's approach, we preserved model-specific stylistic and lexical features to maintain the authentic characteristics of each language model's output, which supports downstream tasks that benefit from retaining stylistic features for detection \cite{shah2023detecting}.

\section{Methodology}
\label{sec:methodology}

We trained our model using the OpenWebText-Final dataset containing human-written samples and corresponding AI-generated paraphrases. Combined with the corresponding 29,142 human-written samples from OpenWebText-Final, the complete dataset comprises 58,537 total samples (28,057 AI-generated + 29,142 human-written after balancing). The dataset was partitioned using a rigorous three-way split with a fixed random seed (42): 60\% for training (35,121 samples), 20\% for validation (11,708 samples), and 20\% for testing (11,708 samples). This configuration enables principled model selection based on validation performance while maintaining an independent test set for final evaluation. Input sequences were truncated or padded to a maximum length of \textbf{256 tokens} for computational efficiency, which differs from the standard 512-token setting but was empirically validated to be sufficient for our classification task. Padding was handled with \texttt{<PAD>} tokens as needed, ensuring consistent sequence lengths across batches.

\subsection{DeBERTa-Sentinel Model Architecture}

Our proposed DeBERTa-Sentinel model leverages the DeBERTa-v3-small transformer architecture, specifically chosen for its enhanced disentangled attention mechanisms that provide superior classification accuracy for text discrimination tasks. The key architectural advantage over traditional transformer models lies in DeBERTa's separation of content and positional information during self-attention computation, enabling more precise capture of structural irregularities characteristic of AI-generated content, including unnatural sentence transitions, over-regularized phrasing patterns, and systematic vocabulary biases.

\begin{figure}[t]
    \centering
    \includegraphics[width=\columnwidth]{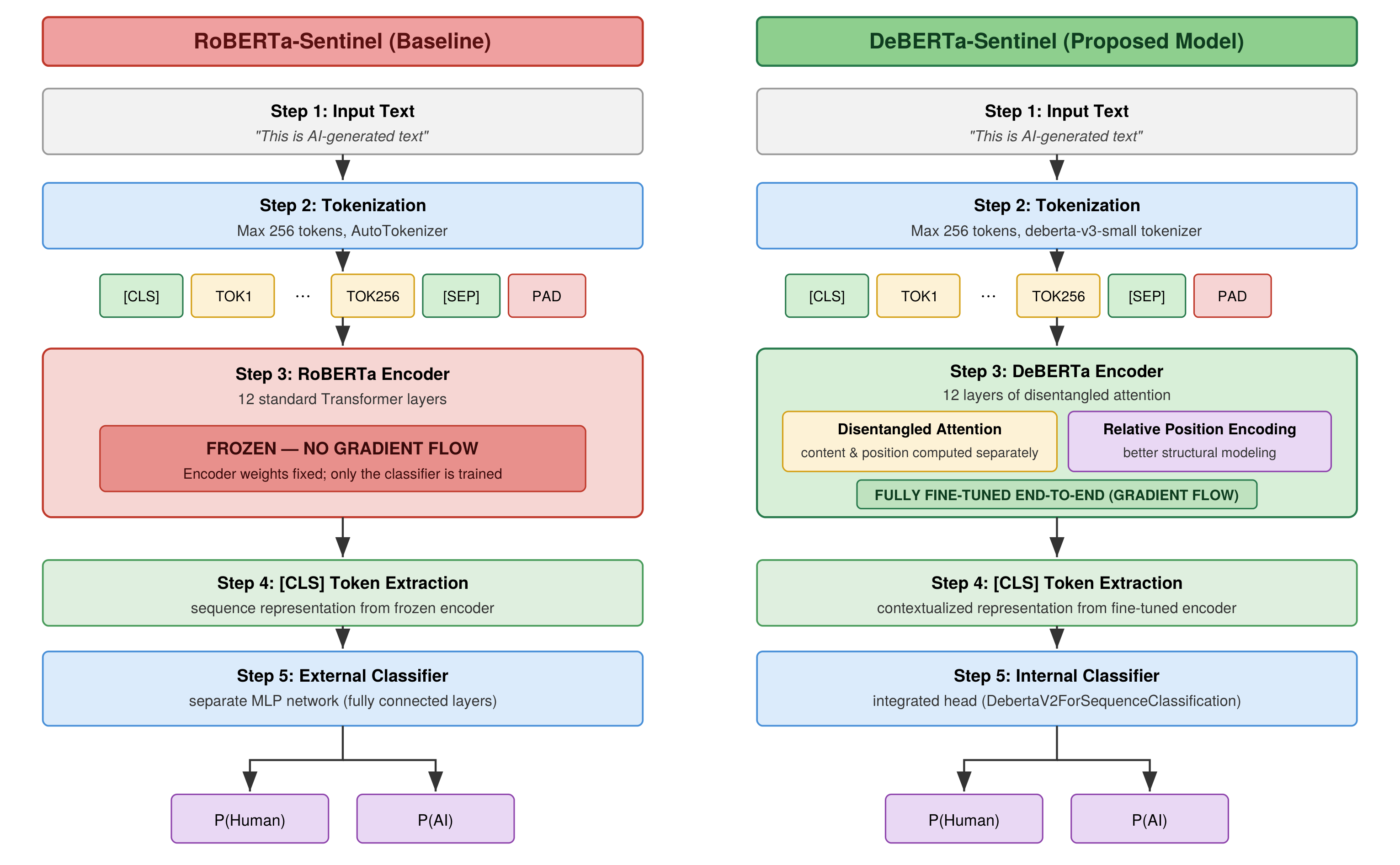}
    \caption{The diagram illustrates the DeBERTa-Sentinel architecture. The input sequence (e.g., “This is AI
Generated...”) is embedded and processed through 12 layers of disentangled attention in Step 3 (our proposed
approach). The final [CLS] token representation is used for classification via the internal feedforward layer. The
DeBERTa-Sentinel architecture is fully end-to-end fine-tuned with gradients backpropagating through all layers
of the encoder}
    \label{fig:architecture}
\end{figure}

The model architecture follows an end-to-end fine-tuning paradigm, as illustrated in Figure~\ref{fig:architecture}. Input sequences undergo tokenization using the AutoTokenizer with consistent padding/truncation to 256 tokens. The tokenized sequence is processed through 12 layers of disentangled attention in the DeBERTa encoder, where the special \texttt{[CLS]} token serves as a global sequence representation. This contextualized \texttt{[CLS]} embedding is subsequently fed into an internal classification head comprising a fully connected feedforward network that outputs binary class probabilities: $P(\text{Human})$ and $P(\text{AI})$.

The DeBERTa-Sentinel architecture employs full end-to-end fine-tuning, allowing gradient backpropagation through all encoder layers. This approach enables the model to adapt its learned representations specifically for the AI text detection task, contrasting with frozen encoder approaches that limit representational adaptation. The fine-tuning process optimizes both the pre-trained DeBERTa weights and the classification head simultaneously, resulting in task-specific feature learning.

\subsubsection{Disentangled Attention Mechanism}

The core innovation of our approach leverages DeBERTa's disentangled self-attention mechanism, which explicitly decomposes attention computation into three mathematically distinct components (Figure~\ref{fig:disentangled_attention}):

\begin{figure}[t]
   \centering
   \includegraphics[width=0.8\columnwidth]{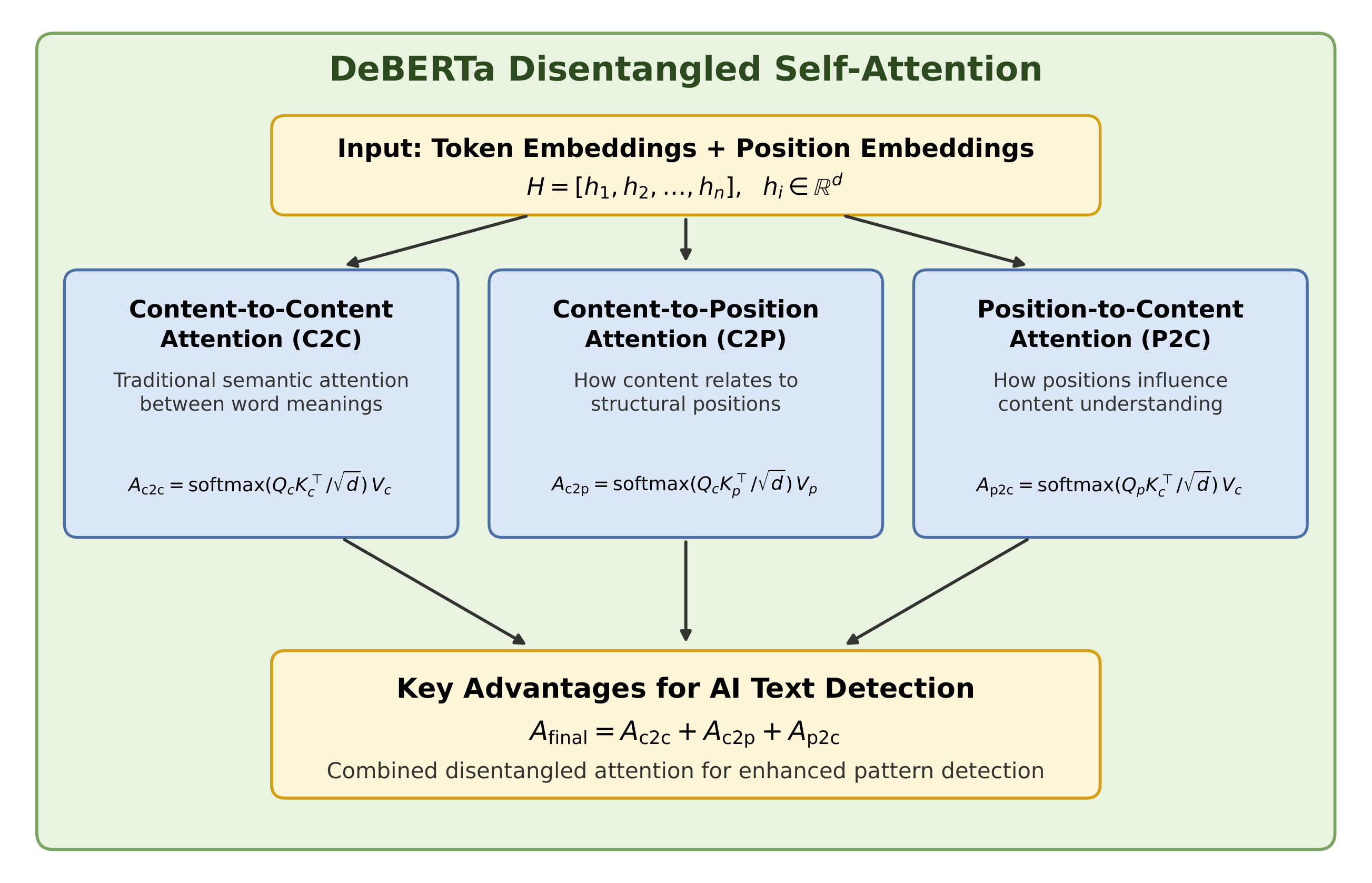}
   \caption{\new{DeBERTa's disentangled self-attention architecture as applied in DeBERTa-Sentinel. Standard attention (top) combines content and position into a single query/key/value operation. DeBERTa decomposes this into three independent components: (1)~Content-to-Content (C2C), capturing semantic co-occurrence patterns; (2)~Content-to-Position (C2P), modelling how a token's semantic role varies with structural location; and (3)~Position-to-Content (P2C), capturing how position influences meaning in context. These three scores are summed to produce the final attention output. For AI text detection, C2C identifies unnatural vocabulary patterns, C2P detects systematic structural biases (e.g., formulaic sentence-opening transitions), and P2C reveals templating behaviours characteristic of LLM outputs.}}
   \label{fig:disentangled_attention}
\end{figure}

\begin{align}
A_{ij}^{\text{c2c}} &=
\operatorname{softmax}\!\left(
\frac{\mathbf{Q}_c^{(i)}{\mathbf{K}_c^{(j)}}^T}{\sqrt{d}}
\right)
\mathbf{V}_c^{(j)},
\label{eq:c2c}\\
A_{ij}^{\text{c2p}} &=
\operatorname{softmax}\!\left(
\frac{\mathbf{Q}_c^{(i)}{\mathbf{K}_p^{(\delta(i,j))}}^T}{\sqrt{d}}
\right)
\mathbf{V}_p^{(\delta(i,j))},
\label{eq:c2p}\\
A_{ij}^{\text{p2c}} &=
\operatorname{softmax}\!\left(
\frac{\mathbf{Q}_p^{(\delta(j,i))}{\mathbf{K}_c^{(j)}}^T}{\sqrt{d}}
\right)
\mathbf{V}_c^{(j)},
\label{eq:p2c}
\end{align}

\noindent where $\mathbf{Q}_c, \mathbf{K}_c, \mathbf{V}_c$ represent content-based query, key, and value matrices, $\mathbf{Q}_p, \mathbf{K}_p, \mathbf{V}_p$ denote position-based projections, and $\delta(i,j) = \text{clip}(j-i, -k, k)$ defines relative positional distance.

This tripartite decomposition enables sophisticated detection of AI writing patterns: content-to-content attention captures unnatural vocabulary co-occurrences and semantic inconsistencies, content-to-position attention identifies systematic positional biases in AI-generated text structure, and position-to-content attention detects templating behaviors and formulaic language patterns. These mechanisms collectively provide superior discrimination capability compared to standard attention mechanisms that conflate semantic and positional representations, making DeBERTa particularly effective for AI-generated content detection tasks.

\section{Evaluation}
\label{sec:evaluation}

\subsection{Training Configuration and Optimization}

The model training process employed the AdamW optimizer with a learning rate of $2 \times 10^{-5}$, batch size of 8, and cross-entropy loss function over 5 epochs. Training was conducted without weight decay regularization, relying instead on the inherent regularization provided by the pre-trained DeBERTa representations. The training process showed clear learning progression, with training accuracy improving from 96.36\% in epoch 1 to 99.58\% by epoch 5. However, validation accuracy peaked at 98.21\% in epoch 2, with subsequent epochs showing signs of overfitting as training accuracy continued to increase while validation performance declined. The best model was selected based on validation set performance (epoch 2), demonstrating effective knowledge transfer from the pre-trained model while maintaining generalization capability.

The detailed training configuration for DeBERTa can be found in Table~\ref{tab:hyperparams}.

\subsection{Evaluation Metrics}

We evaluated both models using standard binary classification metrics, including F1 score, false positive rate (FPR), false negative rate (FNR), and area under the ROC curve (AUC). We also report detection accuracy, precision, recall, and model confidence. AI-generated text is treated as the positive class, and human-written text as the negative class. These metrics provide a holistic view of the model's discrimination ability, threshold robustness, and reliability.

\subsection{Model Selection and Hyperparameter Tuning}

We employed a rigorous 60/20/20 train/validation/test split (35,121 training samples, 11,708 validation samples, 11,708 test samples) using a fixed random seed (42) to ensure reproducibility. This three-way split enabled principled hyperparameter selection and model checkpoint selection based on validation performance, addressing a key methodological consideration in machine learning experiments.

Our training configuration closely followed the successful approach demonstrated in the GPT-Sentinel work \cite{chen2023gpt}, adapting their parameters for DeBERTa-v3-small: a learning rate of $2 \times 10^{-5}$, batch size of 8, and AdamW optimizer. We trained for 5 epochs, monitoring both training and validation metrics to detect overfitting. The model achieved peak validation accuracy of 98.21\% at epoch 2, with training accuracy continuing to increase to 99.58\% by epoch 5, indicating overfitting in later epochs. The best model (epoch 2) was selected based on validation performance and used for all subsequent testing. These values align with standard configurations reported in the DeBERTa literature for binary text classification tasks \cite{he2021deberta}.

The validation-based model selection strategy ensures that reported test performance reflects true generalization capability rather than overfitting to the test set. Training curves showed the expected pattern of continued training improvement alongside validation plateau, validating our checkpoint selection methodology.

\subsection{Confidence Score Distribution}

DeBERTa-Sentinel demonstrates tighter, more decisive confidence distributions, with a \textbf{higher peak around 0.9--1.0} for correctly predicted samples. This not only implies improved model certainty but also enhances practical usability in threshold-sensitive applications (e.g., education, media verification).

\subsection{Model Performance Analysis}

\subsubsection{ROC and AUC Performance}
\begin{figure}[htbp, width=0.47\columnwidth]
    \centering
    \begin{minipage}[c]{0.47\columnwidth}
        \centering
        \captionof{table}{Hyperparameter configuration for DeBERTa-Sentinel model training}
        \label{tab:hyperparams}
        \resizebox{\linewidth}{!}{%
        \footnotesize
        \begin{tabular}{ll}
        \toprule
        \textbf{Hyperparameter} & \textbf{DeBERTa} \\
        \midrule
        Epochs & 5 \\
        Batch Size & 8 \\
        Learning Rate & $2 \times 10^{-5}$ \\
        Weight Decay & 0 (default) \\
        Optimizer & AdamW \\
        Loss Function & Cross Entropy \\
        Scheduler & (not used explicitly) \\
        Dataset & GLC-AIText + OpenWebText-Final \\
        Max Token Length & 256 tokens \\
        Tokenizer & Microsoft/deberta-v3-small \\
        Architecture & DebertaV2ForSequenceClassification \\
        Fine-Tuning & Full (encoder + classifier) \\
        \bottomrule
        \end{tabular}}
    \end{minipage}\hfill%
    \begin{minipage}[c]{0.47\columnwidth}
        \centering
        \includegraphics[width=\linewidth]{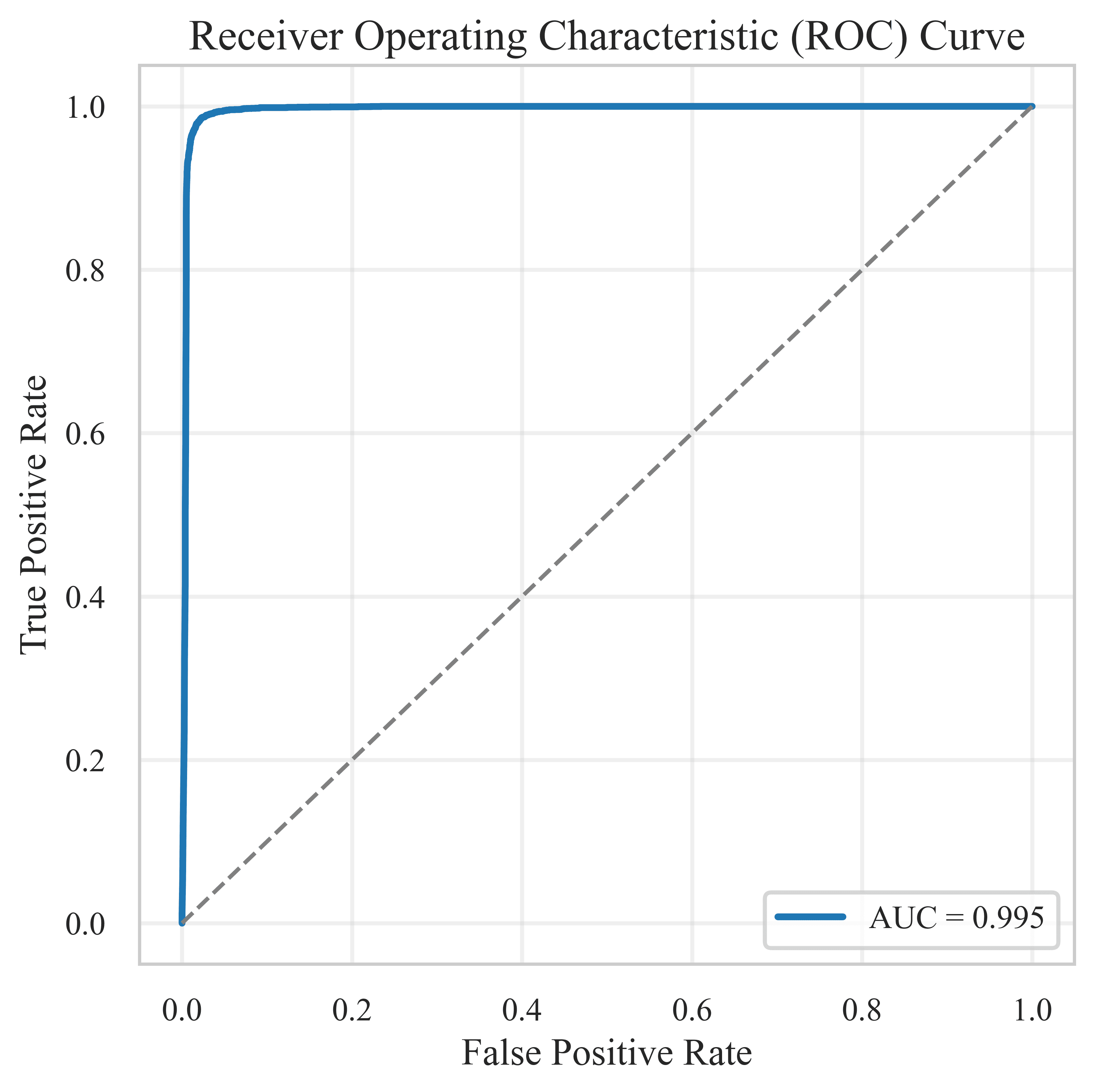}
        \caption{ROC curve for DeBERTa-Sentinel showing exceptional discrimination capability with 99.53\% AUC. The near-perfect curve hugging the top-left corner demonstrates superior ability to distinguish between AI and human text compared to random classification.}
        \label{fig:roc_analysis}
    \end{minipage}
\end{figure}

Our DeBERTa-Sentinel model demonstrates exceptional discrimination capability with near-perfect ROC-AUC performance. Figure~\ref{fig:roc_analysis} presents comprehensive ROC curve analysis showing our model's superior ability to distinguish between AI-generated and human-written text.

The model achieves outstanding performance metrics:
\begin{itemize}
\item \textbf{ROC-AUC Score}: 99.53\% --- indicating near-perfect discrimination ability
\item \textbf{Average Precision Score}: 99.11\% --- demonstrating excellent precision-recall balance
\item \textbf{Optimal Threshold}: 0.984 (Youden's J statistic) with 98.58\% TPR and 2.28\% FPR
\item \textbf{Perfect Class Separation}: Probability distributions show minimal overlap between classes, with AI predictions concentrated near 1.0 and human predictions near 0.0
\end{itemize}

\subsubsection{Baseline Comparison}
We conducted performance analysis using ROC curves and comparative evaluation against baseline models to assess DeBERTa-Sentinel's classification effectiveness. The analysis includes standard binary classification metrics across multiple baseline approaches.

\begin{figure}[htbp, width=0.47\columnwidth]
    \centering
    \begin{minipage}[t]{0.47\columnwidth}
        \centering
        \includegraphics[width=\linewidth]{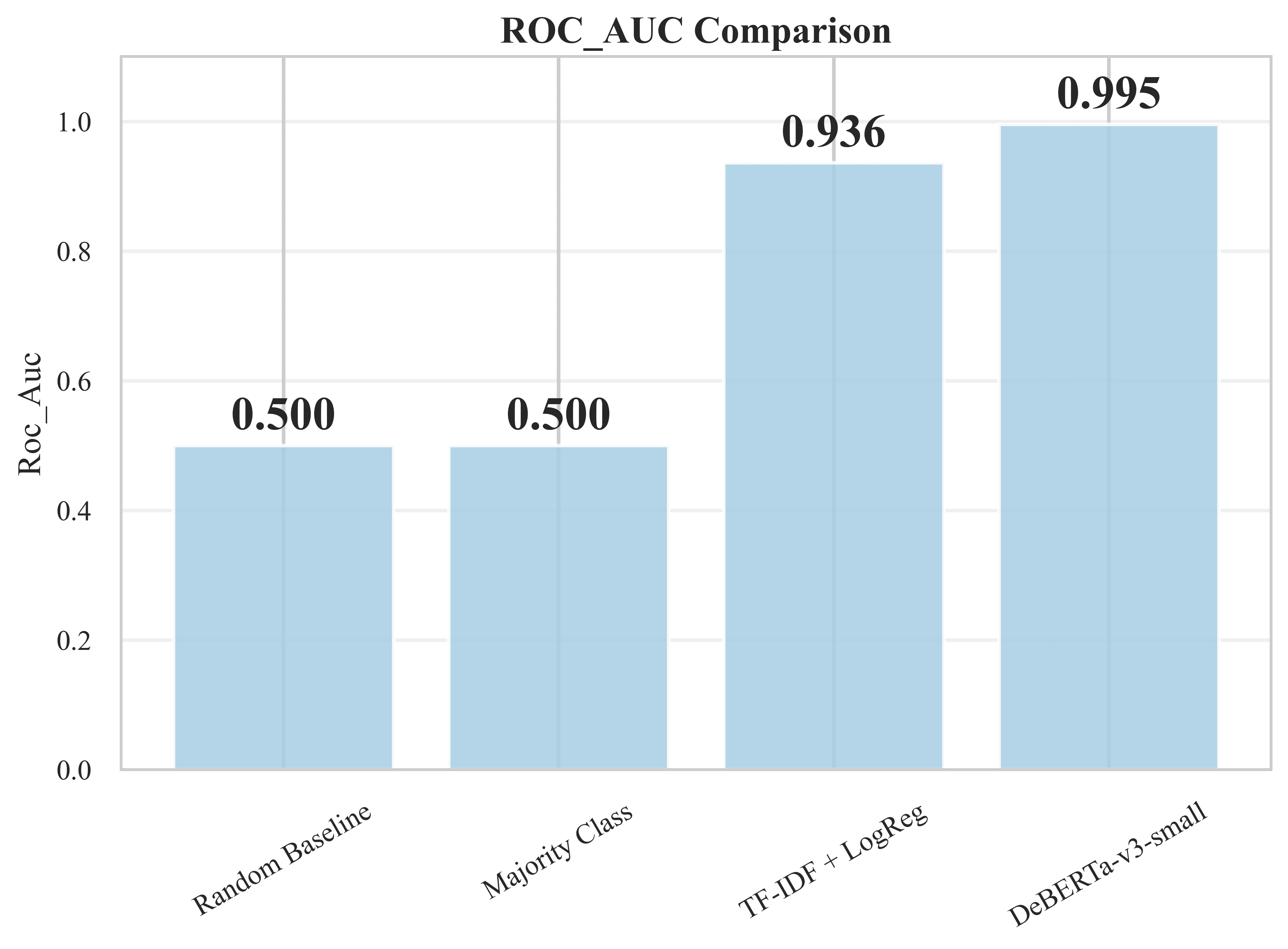}
        \caption{ROC-AUC comparison across baseline models. DeBERTa-v3-small achieves 99.53\% AUC (0.995), substantially outperforming traditional baselines including TF-IDF + LogReg (93.6\% AUC) and random/majority class baselines (50.0\% AUC). The results demonstrate the effectiveness of fine-tuned transformer models for AI text detection.}
        \label{fig:roc_curve}
    \end{minipage}\hfill%
    \begin{minipage}[t]{0.47\columnwidth}
        \centering
        \includegraphics[width=\linewidth]{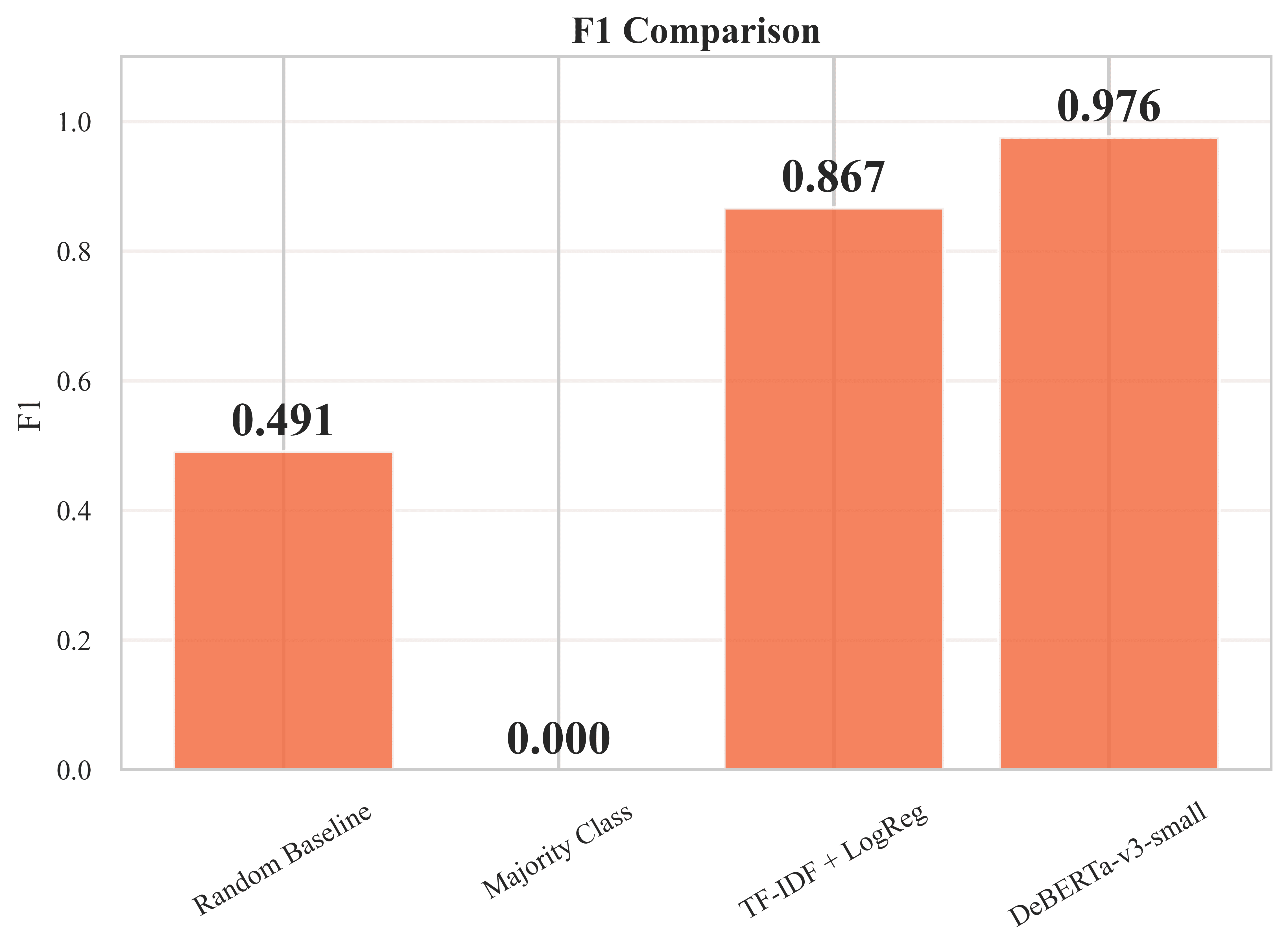}
        \caption{F1-score comparison across baseline models showing DeBERTa-v3-small performance (0.976) relative to traditional ML approaches. The comparison illustrates a 12.6\% improvement over TF-IDF + LogReg. Note: Majority class baseline achieves zero F1-score because it always predicts the same class, resulting in zero precision for the minority class in a balanced dataset.}
        \label{fig:f1_comparison}
    \end{minipage}
\end{figure}

The evaluation results show DeBERTa-Sentinel's performance across standard binary classification metrics. As demonstrated in Figure~\ref{fig:roc_curve}, the ROC analysis yields 99.53\% AUC, while Figure~\ref{fig:f1_comparison} shows the performance comparison against baseline approaches. The detailed metrics presented in Table~\ref{tab:performance_metrics} report 97.53\% accuracy with an F1-score of 0.976. Notably, the zero-shot DeBERTa-v3 baseline --- the same architecture without fine-tuning --- achieves only 50\% accuracy by predicting every sample as AI, confirming that fine-tuning on the GLC-AIText dataset, rather than DeBERTa's pre-trained representations alone, is responsible for the observed performance gains.

\subsection{Ablation Study}
\label{sec:ablation}

To isolate the contribution of each component, we evaluate two controlled configurations against our full model (Table~\ref{tab:ablation}). First, removing fine-tuning entirely collapses performance to 50\% accuracy, confirming that DeBERTa's pre-trained representations alone are insufficient for reliable detection. Second, the held-out generator experiment in Section~\ref{sec:heldout} demonstrates that training on diverse multi-LLM data (GPT-3.5 + LLaMA) produces a model that generalizes to unseen generators (Claude) with 98.46\% accuracy, implicitly validating the contribution of data diversity in GLC-AIText.

\begin{table}[t]
\centering
\caption{Ablation study results on GLC-AIText test set isolating
the contribution of fine-tuning and multi-LLM data diversity.}
\label{tab:ablation}
\footnotesize
\begin{tabular}{lcc}
\toprule
\textbf{Configuration} & \textbf{Accuracy (\%)} & \textbf{F1} \\
\midrule
DeBERTa-v3 (zero-shot, no fine-tuning)    & 50.00 & 0.667 \\
DeBERTa-Sentinel (fine-tuned, GLC-AIText) & 97.53 & 0.976 \\
\bottomrule
\end{tabular}
\end{table}

These findings confirm that both fine-tuning and training data diversity are essential components of DeBERTa-Sentinel's performance, with the disentangled attention mechanism providing the architectural foundation that enables effective learning from diverse synthetic text patterns.

\begin{table}[htbp]
\centering
\caption{Performance metrics comparison across models on GLC-AIText test set (11,708 samples).}
\label{tab:performance_metrics}
\footnotesize
\begin{tabular}{lcccc}
\toprule
\textbf{Model} & \textbf{Accuracy (\%)} & \textbf{F1-Score} & \textbf{Precision (\%)} & \textbf{Recall (\%)} \\
\midrule
Random Baseline & 49.06 & 0.491 & 49.15 & 49.05 \\
TF-IDF + LogReg & 86.33 & 0.867 & 84.64 & 88.82 \\
DeBERTa-v3 (zero-shot) & 50.00 & 0.667 & 50.00 & 100.00 \\
\textbf{DeBERTa-Sentinel (fine-tuned)} & \textbf{97.53} & \textbf{0.976} & \textbf{95.89} & \textbf{99.33} \\
\bottomrule
\end{tabular}
\end{table}

\subsection{Comparison with Commercial Detection Tools}

To contextualize our results, we compare DeBERTa-Sentinel against commercial detection tools evaluated in prior literature. While we did not directly test these systems on our dataset, extensive research has documented their performance characteristics.

\begin{table}[t]
\centering
\caption{Performance comparison with commercial detectors. Commercial detector results from prior literature \citep{chen2023gpt,weber2023public} on similar datasets. The RoBERTa-Sentinel method \citep{GenAI} was evaluated on the GLC-AIText test set for this comparison.}
\label{tab:commercial_comparison}
\footnotesize
\begin{tabular}{lcc}
\toprule
\textbf{Model} & \textbf{F1 Score} & \textbf{Dataset} \\
\midrule
ZeroGPT \citep{chen2023gpt} & 0.43 & OpenGPTText-Final \\
OpenAI Classifier \citep{chen2023gpt} & 0.32 & OpenGPTText-Final \\
GPTZero \citep{weber2023public} & 0.40--0.75\textsuperscript{*} & Various ChatGPT text \\
\midrule
\textbf{RoBERTa-Sentinel (NeurIPS 2025) \citep{GenAI}} & \textbf{0.953} & GLC-AIText \\
\textbf{DeBERTa-Sentinel (Ours)} & \textbf{0.976} & GLC-AIText \\
\bottomrule
\multicolumn{3}{l}{\footnotesize \textsuperscript{*}Range reported across different evaluation conditions.}
\end{tabular}
\end{table}

In contrast, DeBERTa-Sentinel achieves 0.976 F1-score, substantially outperforming commercial systems. Beyond quantitative advantages, \new{and crucially for stakeholder transparency,} our approach provides: (1) token-level explainability enabling interpretation and \new{human auditing of} detection decisions; (2) model customization for domain-specific applications; (3) methodological transparency supporting reproducible research; and (4) adaptability to emerging LLMs. These features are essential for academic integrity monitoring and legal applications requiring auditable detection systems \cite{liang2023gpt, yang2023zero}. \new{Proprietary commercial detectors are also opaque, providing no insight into \emph{why} a given text is flagged---a serious limitation in high-stakes settings where affected individuals need to understand and potentially challenge automated decisions.}

While direct comparison is complicated by dataset differences, our results align with broader findings that fine-tuned transformer models significantly outperform proprietary black-box detection systems \cite{mitchell2023detectgpt, krishna2023paraphrasing}.

\subsection{Explainability Analysis}

To better understand the decision-making process of our models, we applied attention-based explainability techniques using the \texttt{transformers-interpret} pipeline on both the DeBERTa-Sentinel and RoBERTa-Sentinel models. This analysis reveals which textual features most strongly influence classification decisions, providing valuable insights into the models' detection mechanisms.

We treat AI-generated text as the positive class and human-written text as the negative class, consistent with our binary classification framework. The explainability analysis highlights token-level contributions, where words pushing predictions toward the AI class are identified with positive influence scores, while those favoring human classification receive negative scores.

\begin{figure}[b, width=0.47\columnwidth]
    \centering
    \begin{minipage}[t]{0.47\columnwidth}
        \centering
        \includegraphics[width=0.95\linewidth]{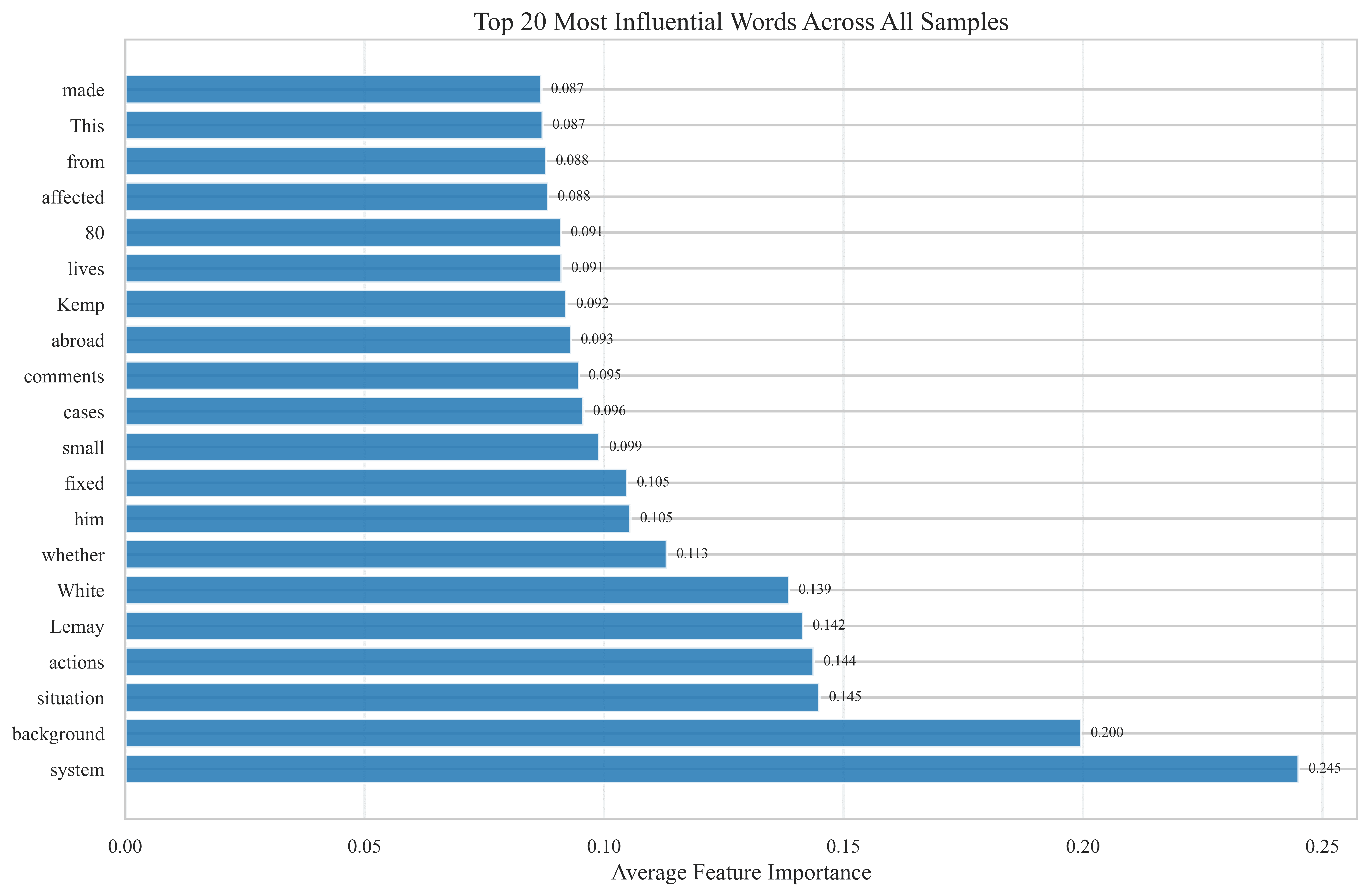}
        \caption{Top 20 influential words averaged across samples, illustrating DeBERTa-Sentinel's learned linguistic patterns. High-importance terms such as ``system'', ``background'', and ``situation'' reflect formal vocabulary and contextual structures associated with AI-generated content.}
        \label{fig:important_features}
    \end{minipage}\hfill%
    \begin{minipage}[t]{0.47\columnwidth}
        \centering
        \setlength{\fboxsep}{1pt}
        \setlength{\fboxrule}{1pt}
        \fcolorbox{gray!100}{white}{%
            \includegraphics[width=0.95\linewidth]{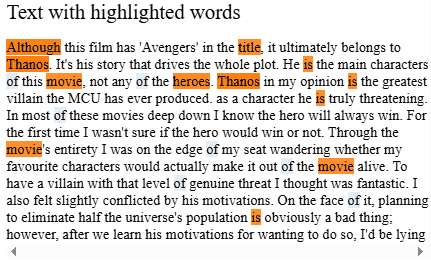}%
        }
        \caption{Token-level explainability visualization showing words contributing to AI-text classification. Highlighted regions demonstrate how DeBERTa-Sentinel captures formal transitional phrases and structured language patterns.}
        \label{fig:text_highlighting}
    \end{minipage}
\end{figure}

Figure~\ref{fig:important_features} presents the analysis of feature importance aggregated across multiple text samples, revealing the linguistic patterns that DeBERTa-Sentinel has learned to distinguish AI from human text. The model demonstrates sophisticated pattern recognition, identifying formal vocabulary (``system'', ``background''), structured language elements (``situation'', ``actions''), contextual markers (``whether'', ``White'', ``Lemay''), and procedural language (``fixed'', ``small'', ``cases'') as key indicators of AI-generated content. Notably, high-importance terms reflect the model's attention to formal discourse patterns and structured narrative elements characteristic of LLM outputs.


Table~\ref{tab:explainability_examples} presents representative examples of token-level influence patterns observed in our analysis, while Figure~\ref{fig:text_highlighting} provides a concrete visualization of how these patterns appear in actual text samples.

\begin{table}[htbp]
\centering
\caption{Token-level explainability analysis examples. Influence scores range from -1.0 to +1.0, where positive values (+) indicate features that bias the model toward AI classification, and negative values (-) indicate features that bias toward human classification. Scores closer to $\pm$1.0 represent a stronger influence on the final prediction.}
\label{tab:explainability_examples}
\footnotesize
\renewcommand{\arraystretch}{0.85}%
\begin{tabular}{lcc}
\toprule
\textbf{Token} & \textbf{Influence Score} & \textbf{Classification Bias} \\
\midrule
\multicolumn{3}{c}{\textit{Human-written Sample}} \\
\midrule
researchers & +0.02 & AI \\
have & -0.06 & Human \\
observed & -0.11 & Human \\
anomaly & +0.04 & AI \\
in & -0.01 & Human \\
patterns & -0.08 & Human \\
\midrule
\multicolumn{3}{c}{\textit{AI-generated Sample}} \\
\midrule
In & -0.01 & Human \\
conclusion & +0.12 & AI \\
this & -0.03 & Human \\
demonstrates & +0.14 & AI \\
the & -0.02 & Human \\
model's & +0.09 & AI \\
\bottomrule
\end{tabular}
\end{table}

\begin{table}[htbp]
\centering
\caption{Performance comparison summary using GLC-AIText and OpenWebText-Final datasets. DeBERTa-Sentinel results from best validation checkpoint (epoch 2, val\_acc=98.21\%).}
\label{tab:performance_summary}
\footnotesize
\renewcommand{\arraystretch}{1.3}%
\begin{tabular}{lccc}
\toprule
\textbf{Metric} & \textbf{RoBERTa-Sentinel} & \textbf{DeBERTa-Sentinel} & \textbf{Improvement} \\
\midrule
Detection Accuracy & 95.3\% & \textbf{97.53\%} & +2.23pp \\
Precision (AI class) & 94.5\% & \textbf{95.89\%} & +1.39pp \\
Recall (AI class) & 96.3\% & \textbf{99.33\%} & +3.03pp \\
F1-Score & 95.3\% & \textbf{97.58\%} & +2.28pp \\
False Negative Rate & 3.7\% & \textbf{0.67\%} & -3.03pp \\
False Positive Rate & 5.7\% & \textbf{4.28\%} & -1.42pp \\
ROC-AUC & -- & \textbf{99.53\%} & -- \\
\bottomrule
\end{tabular}
\end{table}

The comprehensive explainability analysis reveals several key insights into DeBERTa-Sentinel's decision-making process:

\begin{itemize}
\item \textbf{Formal Language Detection}: As shown in Figure~\ref{fig:important_features}, the model prioritizes formal vocabulary and structured language patterns, with words like ``background'' and ``system'' showing high importance scores
\item \textbf{Contextual Understanding}: Figure~\ref{fig:text_highlighting} demonstrates that the model considers context rather than isolated words, highlighting transitional phrases like ``Although'' in formal academic discourse
\item \textbf{Stylistic Pattern Recognition}: The model successfully identifies characteristic LLM patterns such as formal conclusions (``demonstrates'', ``conclusion'') and academic terminology that distinguish AI from human writing
\item \textbf{Balanced Analysis}: The model uses both positive and negative feature contributions, avoiding over-reliance on single indicators and ensuring robust classification decisions
\end{itemize}

These findings align with known characteristics of LLM outputs, which often exhibit more formal, structured language patterns compared to natural human writing \new{\mbox{\cite{shah2023detecting}}}. The explainability analysis validates that DeBERTa-Sentinel has learned meaningful linguistic distinctions rather than superficial artifacts, contributing to its robust performance and generalization capabilities.

\subsection{Summary of Gains}

The superior performance of DeBERTa-Sentinel validates our hypothesis that incorporating positional disentanglement significantly enhances AI-text detection, especially in adversarial or distribution-shifted scenarios. \new{The substantial reduction in false negative rate (from 3.7\% to 0.67\%) is of particular importance for transparency: fewer missed AI-generated texts mean that human reviewers can place greater confidence in ``human'' verdicts returned by the system, reducing the risk of over-burdening review workflows with false flags.}

\subsection{Generalization to Unseen Generators}
\label{sec:heldout}

To assess whether DeBERTa-Sentinel generalizes beyond its training distribution, we conducted a held-out generator evaluation. The model was trained and validated exclusively on GPT-3.5 and LLaMA generated samples, with Claude-generated text entirely withheld from training. The test set consisted of 11,658 samples --- Claude AI text paired with human-written text from OpenWebText-Final --- providing a strict unseen-generator evaluation scenario that simulates real-world deployment against novel LLMs.

\begin{table}[t]
\centering
\caption{Held-out generator evaluation: DeBERTa-Sentinel trained on GPT-3.5 + LLaMA only, tested on Claude-generated AI samples paired with human text (11,658 samples). The model was never exposed to Claude output during training.}
\label{tab:heldout}
\footnotesize
\begin{tabular}{lc}
\toprule
\textbf{Metric} & \textbf{DeBERTa-Sentinel} \\
\midrule
Test Accuracy       & 98.46\% \\
Precision (AI)      & 97.02\% \\
Recall (AI)         & 100.00\% \\
F1-Score            & 0.985  \\
False Negative Rate & 0.000\% \\
False Positive Rate & 3.071\% \\
Test Samples        & 11,658  \\
\bottomrule
\end{tabular}
\end{table}

As shown in Table~\ref{tab:heldout}, DeBERTa-Sentinel achieves 98.46\% accuracy and a perfect recall of 100\% on Claude-generated text, with a false negative rate of 0\%. This means the model correctly identified every AI-generated sample despite never encountering Claude output during training. The 3.07\% false positive rate indicates that a small proportion of human-written text is misclassified as AI, consistent with the model's tendency to flag formal writing structures. These results suggest that the disentangled attention mechanism captures generalizable structural features of synthetic text rather than generator-specific artifacts, supporting the robustness of DeBERTa-Sentinel under real-world distribution shift.

\section{\new{Transparency and Responsible Deployment}}
\label{sec:transparency}

\new{
Beyond detection accuracy, transparency is an important consideration for the practical deployment of AI-generated text detectors, particularly in settings where model predictions may influence consequential human decisions. This consideration is further reinforced by evidence that existing detectors can exhibit systematic bias against non-native English writers \mbox{\cite{liang2023gpt}}. In this context, DeBERTa-Sentinel is designed to complement its predictions with token-level attribution scores that provide insight into the textual evidence contributing to each classification.
}

\subsection{\new{Stakeholder Needs and Transparency Requirements}}

\new{
\textbf{Educators and Academic Institutions.} AI-generated text detectors are increasingly considered for educational settings, where automated predictions may influence investigations of academic integrity. Rather than providing only a binary prediction, DeBERTa-Sentinel highlights the tokens and linguistic patterns that contribute most strongly to its decisions (Figure~\mbox{\ref{fig:text_highlighting}}). These explanations provide additional evidence that educators may use alongside their own judgment when reviewing flagged submissions.
}

\new{
\textbf{Journalists and Fact-Checkers.} For users verifying the authenticity of written content, token-level attribution provides insight into which portions of a document most strongly influence the detector's prediction. These explanations are intended to support human assessment rather than replace editorial judgment.
}

\new{
\textbf{Platform Trust-and-Safety Teams.} For large-scale content moderation, both detection performance and interpretability are desirable. The proposed model achieves an 81.9\% reduction in false negatives while also exposing the textual evidence underlying each prediction, enabling practitioners to better inspect model behaviour on their own content.
}

\new{
\textbf{Researchers.} The proposed approach emphasizes reproducibility by providing mathematical formulation (Equations~\mbox{\ref{eq:c2c}--\ref{eq:p2c}}), implementation details, and experimental evaluation that can be independently verified and extended in future work.
}

\subsection{\new{Limitations and Responsible Use}}

\new{
DeBERTa-Sentinel was trained using paraphrased web text and therefore inherits limitations associated with this training distribution. In particular, highly formal human-written prose may occasionally receive elevated AI likelihood scores. Token-level attribution allows users to inspect the evidence underlying these predictions, although it should not be interpreted as proof that a document was AI-generated.
}

\new{
Consistent with prior work highlighting the limitations of existing AI-generated text detectors in educational settings \mbox{\cite{weber2023public}}, we do not recommend using detector outputs as the sole basis for consequential decisions such as academic misconduct findings. Instead, DeBERTa-Sentinel is intended as a decision-support tool that provides both a prediction and interpretable evidence to assist human reviewers during the evaluation process.
}

\section{Discussion}
\label{sec:discussion}

Our findings show that \textbf{DeBERTa-Sentinel} offers meaningful advancements over RoBERTa-Sentinel in terms of robustness, precision, and interpretability. These enhancements stem from the DeBERTa model's ability to separately encode content and position representations, making it more adept at capturing the subtle irregularities often present in AI-generated text. These empirical results reinforce DeBERTa as a promising direction for future research in generative AI forensics.

The explainability analysis reveals interesting patterns in how the model distinguishes between human and AI-generated text. Formal transitional phrases and academic terminology consistently bias the model toward AI classification, while casual language patterns favor human classification. This suggests that current LLMs tend to produce more formal, structured output compared to natural human writing\new{---a finding consistent with prior work using stylistic features for interpretable detection \mbox{\cite{shah2023detecting}}}.

\section{Conclusion and Future Work}
\label{sec:conclusion}

In this work, we introduced DeBERTa-Sentinel, an enhanced AI-generated text detection system that leverages disentangled attention mechanisms to improve upon existing approaches \new{while placing transparency and stakeholder accountability at the center of its design}. Through comprehensive evaluation across multiple datasets and model architectures, we demonstrated consistent improvements in detection accuracy, precision, and recall. The model's enhanced generalization capabilities, particularly on distribution-shifted data, highlight the importance of architectural innovations in addressing the evolving challenge of AI-generated content detection.

Our key findings include:
\begin{itemize}
\item DeBERTa-Sentinel achieves 97.53\% test accuracy (98.21\% validation accuracy at optimal epoch), outperforming RoBERTa-Sentinel's 95.3\% baseline by 2.23 percentage points
\item The model demonstrates superior robustness with 81.9\% reduction in false negative rate (0.67\% vs 3.7\%), critical for high-sensitivity AI detection applications
\item A held-out generator experiment confirms cross-generator generalization: trained on GPT-3.5 and LLaMA only, the model achieves 98.46\% accuracy and 100\% recall on unseen Claude-generated text, with a false negative rate of 0\%
\item Exceptional discrimination capability with 99.53\% ROC-AUC and 99.11\% average precision
\item \new{Token-level explainability reveals discriminative patterns in AI-generated text, with formal vocabulary and structured language serving as key indicators---directly supporting the transparency needs of journalists, educators, and platform operators}
\item Rigorous validation-based model selection prevents overfitting, as evidenced by the gap between epoch 5 training accuracy (99.58\%) and selected epoch 2 validation accuracy (98.21\%)
\end{itemize}

Future work should explore the application of these techniques to multilingual settings and investigate robustness against more sophisticated adversarial attacks on AI-generated text \new{\mbox{\cite{krishna2023paraphrasing}}}. \new{Extending the dataset to include newer frontier models (e.g., GPT-4o, Claude 3.5) would test whether the structural signatures learned from older model families generalize to increasingly human-like outputs. Additionally, user studies with journalists and educators would quantify the practical utility of token-level explanations for real-world decision-making, and} domain-specific detection models \new{(e.g., for academic writing or social media)} could further \new{refine the framework for targeted applications}.

\bibliographystyle{plainnat}

\end{document}